\pdfoutput=1
\documentclass[11pt,a4paper]{article}
\usepackage[hyperref]{acl2020}
\usepackage{times}
\usepackage{latexsym}

\usepackage{microtype}
\usepackage{tabularx}

\usepackage{tikz}
\usepackage{amsmath}
\usepackage{amssymb}
\usepackage{booktabs}
\usepackage{multirow}
\usepackage{url}
\usepackage{tikz}
\usetikzlibrary{positioning, arrows, calc}

\newcommand{\bftab}{\fontseries{b}\selectfont}
\renewcommand{\vec}{\mathbf}
\newcommand{\narrow}{\setlength{\tabcolsep}{1.4mm}}
\newcommand{\verynarrow}{\setlength{\tabcolsep}{.9mm}}

\aclfinalcopy 
\def\aclpaperid{1723} 

\title{Sentence Meta-Embeddings for Unsupervised Semantic Textual Similarity}

\author{Nina Poerner$^{\ast\dagger}$ and Ulli Waltinger$^\dagger$ and Hinrich Sch{\"u}tze$^\ast$ \\
  $^\ast$Center for Information and Language Processing, LMU Munich, Germany \\
  $^\dagger$Corporate Technology Machine Intelligence (MIC-DE), Siemens AG Munich, Germany \\
{\tt poerner@cis.uni-muenchen.de | inquiries@cislmu.org} }

\date{}

\begin{document}
\maketitle
\begin{abstract}
We address the task of unsupervised Semantic Textual Similarity (STS) by ensembling diverse pre-trained sentence encoders into \textit{sentence meta-embeddings}.
We apply, extend and evaluate different meta-embedding methods from the word embedding literature at the sentence level, including dimensionality reduction \cite{yin2016learning}, generalized Canonical Correlation Analysis \cite{rastogi2015multiview} and cross-view auto-encoders \cite{bollegala2018learning}.
Our sentence meta-embeddings set a new unsupervised State of The Art (SoTA) on the STS Benchmark and on the STS12--STS16 datasets, with gains of between 3.7\% and 6.4\% Pearson's $r$ over single-source systems.
\end{abstract}

\section{Introduction}
Word meta-embeddings have been shown to exceed single-source word embeddings on word-level semantic benchmarks \cite{yin2016learning, bollegala2018learning}.
Presumably, this is because they combine the complementary strengths of their components.

There has been recent interest in pre-trained ``universal'' sentence encoders, i.e., functions that encode diverse semantic features of sentences into fixed-size vectors \cite{conneau2017supervised}.
Since these sentence encoders differ in terms of their architecture and training data, we hypothesize that their strengths are also complementary and that they can benefit from meta-embeddings.

To test this hypothesis, we adapt different meta-embedding methods from the word embedding literature.
These include dimensionality reduction \cite{yin2016learning}, cross-view autoencoders \cite{bollegala2018learning} and Generalized Canonical Correlation Analysis (GCCA) \cite{rastogi2015multiview}.
The latter method was also used by \newcite{poerner2019multiview} for domain-specific Duplicate Question Detection.

Our sentence encoder ensemble includes three models from the recent literature: Sentence-BERT \cite{reimers2019sentence}, the Universal Sentence Encoder \cite{cer2017semeval} and averaged ParaNMT vectors \cite{wieting2018paranmt}.
Our meta-embeddings outperform every one of their constituent single-source embeddings on STS12--16 \cite{agirre2016semeval} and on the STS Benchmark \cite{cer2017semeval}.
Crucially, since our meta-embeddings are agnostic to the contents of their ensemble, future improvements may be possible by adding new encoders.

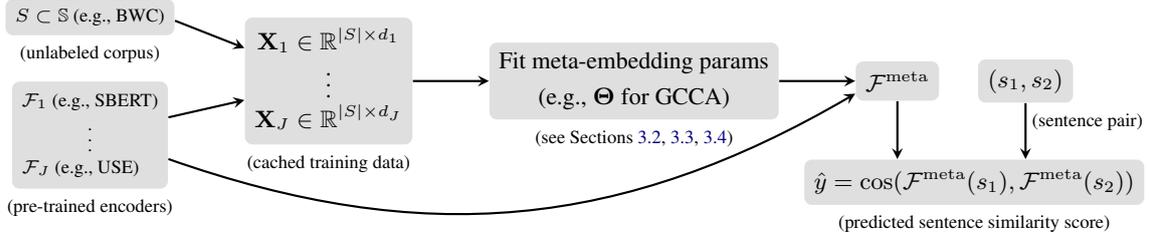
\begin{figure*}
\centering
\begin{tikzpicture}
\node [rectangle, fill=gray!25, rounded corners=1mm] at (0,0) (encoders) {\scriptsize $\begin{matrix}\mathcal{F}_1 \text{ (e.g., SBERT)} \\ \vdots \\ \mathcal{F}_J \text{ (e.g., USE)\phantom{....}}\end{matrix}$};
\node [below=0pt of encoders] (encoderslabel) {\scriptsize (pre-trained encoders)};
\node [rectangle, fill=gray!25, rounded corners=1mm, above=6mm of encoders] (crp) {\scriptsize $S \subset \mathbb{S}$ (e.g., BWC)};
\node [below=0pt of crp] (crplabel) {\scriptsize (unlabeled corpus)};
\node [rectangle, fill=gray!25, rounded corners=1mm, right=1cm of encoders.north east, align=left] (inputs) {\small $\begin{matrix} \mathbf{X}_1\in \mathbb{R}^{|S|\times d_1} \\ \vdots \\ \mathbf{X}_J\in \mathbb{R}^{|S|\times d_J} \end{matrix}$};
\node [below=0pt of inputs] (inputslabel) {\scriptsize (cached training data)};
\draw [->, >=stealth, thick] (encoders) -- (inputs);
\draw [->, >=stealth, thick] ([yshift=2mm] crp.south east) -- (inputs);
\node [rectangle, fill=gray!25, rounded corners=1mm, right=1cm of inputs, align=center] (metafit) {\small Fit meta-embedding params \\ \small (e.g., $\mathbf{\Theta}$ for GCCA)};
\node [below=0pt of metafit] (metafitlabel) {\scriptsize (see Sections \ref{sec:svd}, \ref{sec:gcca}, \ref{sec:ae})};
\draw [->, >=stealth, thick] (inputs) -- (metafit);
\node [rectangle, fill=gray!25, rounded corners=1mm, right=1cm of metafit] (meta) {\small $\mathcal{F}^\mathrm{meta}$};
\draw [->, >=stealth, thick] (metafit) -- (meta);
\draw [->, >=stealth, thick, bend angle=25, rounded corners=1mm, bend right] ([yshift=4mm] encoders.south east) to ([yshift=1mm] meta.south west);
\node [right=5mm of meta, rectangle, fill=gray!25, rounded corners=1mm] (test) {\small $(s_1, s_2)$};
\node [below=0mm of test.south east] (testlabel) {\scriptsize \phantom{......}(sentence pair)};
\node at ($(meta)!0.6!(test)$) (tmp) {};
\node [rectangle, fill=gray!25, rounded corners=1mm, below=9mm of tmp] (pred) {\small $\hat{y} = \mathrm{cos}(\mathcal{F}^\mathrm{meta}(s_1), \mathcal{F}^\mathrm{meta}(s_2))$};
\node [below=0pt of pred] (predlabel) {\scriptsize (predicted sentence similarity score)};
\draw [->, >=stealth, thick] (test.south) to (pred.north-|test.south);
\draw [->, >=stealth, thick] (meta.south) to (pred.north-|meta.south);
\end{tikzpicture}
\caption{Schematic depiction: Trainable sentence meta-embeddings for unsupervised STS.}
\end{figure*}

\section{Related work}
\subsection{Word meta-embeddings}
Word embeddings are functions that map word types to vectors.
They are typically trained on unlabeled corpora and capture word semantics (e.g., \citet{mikolov2013distributed, pennington2014glove}).
 
Word meta-embeddings combine ensembles of word embeddings by various operations:
\citet{yin2016learning} use concatenation, SVD and linear projection,
\citet{coates2018frustratingly} show that averaging word
embeddings has properties similar to concatenation.
\citet{rastogi2015multiview} apply generalized canonical
correlation analysis (GCCA) to an ensemble of word vectors.
\citet{bollegala2018learning} learn word meta-embeddings using autoencoder architectures.
\citet{neill2018angular} evaluate different loss functions for autoencoder word meta-embeddings, while \citet{bollegala2018think} explore locally linear mappings.

\subsection{Sentence embeddings}
Sentence embeddings are methods that produce one vector per sentence.
They can be grouped into two categories:

(a) Word embedding average sentence encoders take a (potentially weighted) average of pre-trained word embeddings.
Despite their inability to understand word order, they are surprisingly effective on sentence similarity tasks \cite{arora2017simple, wieting2018paranmt, ethayarajh2018unsupervised}

(b) Complex contextualized sentence encoders, such as Long Short Term Memory Networks (LSTM) \cite{hochreiter1997long} or Transformers \cite{vaswani2017attention}.
Contextualized encoders can be pre-trained as unsupervised language models \cite{peters2018deep, devlin2019bert}, but they are usually improved on supervised transfer tasks such as Natural Language Inference \cite{bowman2015large}.

\subsection{Sentence meta-embeddings} 
Sentence meta-embeddings have been explored less frequently than their word-level counterparts.
\citet{kiela2018dynamic} create meta-embeddings by training an LSTM sentence encoder on top of a set of dynamically combined word embeddings.
Since this approach requires labeled data, it is not applicable to unsupervised STS.

\citet{tang2019improving} train a Recurrent Neural Network (RNN) and a word embedding average encoder jointly on a large corpus to predict similar representations for neighboring sentences.
Their approach trains both encoders from scratch, i.e., it cannot be used to combine existing encoders.

\citet{poerner2019multiview} propose a GCCA-based multi-view sentence encoder that combines domain-specific and generic sentence embeddings for unsupervised Duplicate Question Detection.
In this paper, we extend their approach by exploring a wider range of meta-embedding methods and an ensemble that is more suited to STS.

\subsection{Semantic Textual Similarity (STS)} 
\label{sec:relatedsts}
Semantic Textual Similarity (STS) is the task of rating the similarity of two natural language sentences on a real-valued scale.
Related applications are semantic search, duplicate detection and sentence clustering.

Supervised SoTA systems for STS typically apply cross-sentence attention \cite{devlin2019bert, raffel2019exploring}.
This means that they do not scale well to many real-world tasks.
Supervised ``siamese'' models \cite{reimers2019sentence} on the other hand, while not competitive with cross-sentence attention, can cache sentence embeddings independently of one another.
For instance, to calculate the pairwise similarities of $N$ sentences, a cross-sentence attention system must calculate $O(N^2)$ slow sentence pair embeddings, while the siamese model calculates $O(N)$ slow sentence embeddings and $O(N^2)$ fast vector similarities.

Our meta-embeddings are also cacheable (and hence scalable), but they do not require supervision.

\section{Sentence meta-embedding methods}
\label{sec:methods}
Below, we assume access to an ensemble of pre-trained sentence encoders, denoted $\mathcal{F}_1 \ldots \mathcal{F}_J$.
Every $\mathcal{F}_j$ maps from the (infinite) set of possible sentences $\mathbb{S}$ to a fixed-size $d_j$-dimensional vector.

Word meta-embeddings are usually learned from a finite vocabulary of word types \cite{yin2016learning}.
Sentence embeddings lack such a ``vocabulary'', as they can encode any member of $\mathbb{S}$.
Therefore, we train on a sample $S \subset \mathbb{S}$, i.e., on a corpus of unlabeled sentences.

\subsection{Naive meta-embeddings}
We create naive sentence meta-embeddings by concatenating \cite{yin2016learning} or averaging\footnote{If embeddings have different dimensionalities, we pad the shorter ones with zeros.} \cite{coates2018frustratingly} sentence embeddings.
$$\mathcal{F}^\mathrm{conc}(s') = \begin{bmatrix}\hat{\mathcal{F}}_1(s') \\ \ldots \\ \hat{\mathcal{F}}_J(s') \end{bmatrix}$$
	$$\mathcal{F}^\mathrm{avg}(s') = \sum_j \frac{\hat{\mathcal{F}}_j(s')}{J}$$
Note that we length-normalize all embeddings to ensure equal weighting:
$$\hat{\mathcal{F}}_j(s) = \frac{{\mathcal{F}}_j(s)}{||{\mathcal{F}}_j(s)||_2}$$

\subsection{SVD}
\label{sec:svd}
\citet{yin2016learning} use Singular Value Decomposition (SVD) to compactify concatenated word embeddings.
The method is straightforward to extend to sentence meta-embeddings.
Let $\vec X^\mathrm{conc} \in \mathbb{R}^{|S| \times \sum_j d_j}$ with 
$$\vec x^\mathrm{conc}_n = \mathcal{F}^\mathrm{conc}(s_n) - \mathbb{E}_{s \in S}[\mathcal{F}^\mathrm{conc}(s)]$$
Let $\vec U \vec S \vec V^T \approx \vec X^\mathrm{conc}$ be the $d$-truncated SVD.
The SVD meta-embedding of a new sentence $s'$ is:
$$
\mathcal{F}^\mathrm{svd}(s') = \vec V^T (\mathcal{F}^\mathrm{conc}(s') - \mathbb{E}_{s\in S}[\mathcal{F}^\mathrm{conc}(s)])
$$

\subsection{GCCA}
\label{sec:gcca}
Given random vectors $\vec x_1, \vec x_2$, Canonical Correlation Analysis (CCA) finds linear projections such that $\boldsymbol \theta_1^T \vec x_1$ and $\boldsymbol \theta_2^T \vec x_2$ are maximally correlated.
Generalized CCA (GCCA) extends CCA to three or more random vectors.
\citet{bach2002kernel} show that a variant of GCCA reduces to a generalized eigenvalue problem on block matrices:
$$
\begin{aligned}
\rho
&
\begin{bmatrix} 
\vec \Sigma_{1,1} & \vec 0 & \vec 0 \\
\vec 0 & \vec \Sigma_{\ldots} & \vec 0 \\
\vec 0 & \vec 0 & \vec \Sigma_{J,J} \\
\end{bmatrix} 
\begin{bmatrix}
\boldsymbol \theta_1 \\
\ldots \\
\boldsymbol \theta_J \\
\end{bmatrix}
\\
= &
\begin{bmatrix} 
\vec 0 & \vec \Sigma_{\ldots} & \vec \Sigma_{1,J} \\
\vec \Sigma_{\ldots} & \vec 0 & \vec \Sigma_{\ldots} \\
\vec \Sigma_{J,1} & \vec \Sigma_{\ldots} & \vec 0 \\
\end{bmatrix} 
\begin{bmatrix}
\boldsymbol \theta_1 \\
\ldots \\
\boldsymbol \theta_J \\
\end{bmatrix}
\end{aligned}
$$
where 
$$\begin{aligned}
\vec \Sigma_{j,j'} & = \mathbb{E}_{s\in S}[(\mathcal{F}_j(s)-\boldsymbol{\mu}_j)(\mathcal{F}_{j'}(s)-\boldsymbol{\mu}_{j'})^T] \\
\boldsymbol{\mu}_j & = \mathbb{E}_{s\in S} [\mathcal{F}_j(s)] \\
\end{aligned}
$$
For stability, we add $\frac{\tau}{d_j} \sum_{n=1}^{d_j} \mathrm{diag}(\vec \Sigma_{j,j})_n$ to  $\mathrm{diag}(\vec \Sigma_{j,j})$, where $\tau$ is a hyperparameter.
We stack the eigenvectors of the top-$d$ eigenvalues into matrices $\vec \Theta_j \in \mathbb{R}^{d \times d_j}$ and define the GCCA meta-embedding of sentence $s'$ as:
$$
\mathcal{F}^\mathrm{gcca}(s') = \sum_{j=1}^J \vec \Theta_j (\mathcal{F}_j(s') - \boldsymbol{\mu}_j)
$$
$\mathcal{F}^\mathrm{gcca}$ corresponds to MV-DASE in \citet{poerner2019multiview}.

\begin{table}
\centering
\footnotesize
\narrow
\begin{tabularx}{.98\columnwidth}{Xr|ccccc}
\toprule
& &\multicolumn{4}{c}{loss function} \\
& & MSE & MAE & KLD & (1-COS)$^2$ \\
\midrule
\multirow{3}{*}{\rotatebox{90}{\parbox{1cm}{\centering number hidden layers}}} & 0 & 83.0/84.2 & 84.2/85.1 & 83.0/84.2 & 82.4/83.5 \\
& 1 & 82.7/83.9 & 83.8/84.6 & {\bftab 85.1}/{\bftab 85.5} & 83.3/83.4  \\
& 2 & 82.5/82.8 & 81.3/82.1 & 83.3/83.4 & 82.3/82.3 \\
\bottomrule
\end{tabularx}
\begin{tabularx}{.98\columnwidth}{ccccc}
\toprule
$\tau=10^{-2}$ & $\tau=10^{-1}$ & $\tau=10^{0}$ & $\tau=10^{1}$ & $\tau=10^{2}$  \\
\midrule
84.2/84.1 & 84.8/84.7 & 85.5/85.7 & {\bftab 85.5}/{\bftab 86.1} & 84.9/85.9   \\
\bottomrule
\end{tabularx}
\caption{Hyperparameter search on STS Benchmark development set for AE (top) and GCCA (bottom). Pearson's $r \times 100$ / Spearman's $\rho \times 100$.}
\label{tab:hyper}
\end{table}

\subsection{Autoencoders (AEs)}
\label{sec:ae}
Autoencoder meta-embeddings are trained by gradient descent to minimize some cross-embedding reconstruction loss.
For example, \citet{bollegala2018learning} train feed-forward networks (FFN) to encode two sets of word embeddings into a shared space, and then reconstruct them such that mean squared error with the original embeddings is minimized.
\newcite{neill2018angular} evaluate different reconstruction loss functions: Mean Squared Error (MSE), Mean Absolute Error (MAE), KL-Divergence (KLD) or squared cosine distance (1-COS)$^2$.

We extend their approach to sentence encoders as follows:
Every sentence encoder $\mathcal{F}_j$ has a trainable encoder $\mathcal{E}_j: \mathbb{R}^{d_j} \rightarrow \mathbb{R}^d $ and a trainable decoder $\mathcal{D}_j: \mathbb{R}^d \rightarrow \mathbb{R}^{d_j}$, where $d$ is a hyperparameter.
Our training objective is to reconstruct every embedding $\vec x_{j'}$ from every $\mathcal{E}_j(\vec x_j)$.
This results in $J^2$ loss terms, which are jointly optimized:
$$L(\vec x_1 \ldots \vec x_J) = \sum_j \sum_{j'} l(\vec x_{j'}, \mathcal{D}_{j'}(\mathcal{E}_j(\vec x_j))) $$
where $l$ is one of the reconstruction loss functions listed above.
The autoencoder meta-embedding of a new sentence $s'$ is: 
$$\mathcal{F}^\mathrm{ae}(s') = \sum_j \mathcal{E}_j(\mathcal{F}_j(s'))$$

\newcommand{\blk}{\phantom{00.}}

\begin{table*}[ht]
\centering
\footnotesize
\narrow
\begin{tabularx}{.98\textwidth}{Xr|cccccc}
\toprule
\multicolumn{2}{r|}{dimensionality} & STS12 & STS13 & STS14 & STS15 & STS16 & STS-B  \\
\midrule \midrule
single:ParaNMT  & $d=600\phantom{0}$ & \underline{67.5}/66.3&62.7/62.8&\underline{77.3}/\underline{74.9}&\underline{80.3}/\underline{80.8}&\underline{78.3}/\underline{79.1}&\underline{79.8}/78.9\\ 
single:USE & $d=512\phantom{0}$ & 62.6/63.8&57.3/57.8&69.5/66.0&74.8/77.1&73.7/76.4&76.2/74.6\\
single:SBERT & $d=1024$ & 66.9/\underline{66.8}&\underline{63.2}/\underline{64.8}&74.2/74.3&77.3/78.3&72.8/75.7&76.2/\underline{79.2} \\
\midrule
single:ParaNMT -- up-projection$^\ast$ & $d=1024$ & 67.3/66.2&62.1/62.4&77.1/74.7&79.7/80.2&77.9/78.7&79.5/78.6  \\
single:USE -- up-projection$^\ast$ & $d=1024$ & 62.4/63.7&57.0/57.5&69.4/65.9&74.7/77.1&73.6/76.3&76.0/74.5  \\
\midrule
\midrule
meta:conc & $d=2136$ &72.7/71.3&68.4/68.6&81.0/79.0&84.1/{\bftab 85.5}&{\bftab 82.0}/{\bftab 83.8}&82.8/83.4\\
meta:avg & $d=1024$ &72.5/71.2&68.1/68.3&80.8/78.8&83.7/85.1&81.9/83.6&82.5/83.2 \\
meta:svd & $d=1024$ & 71.9/70.8&68.3/68.3&80.6/78.6&83.8/85.1&81.6/83.6&83.4/83.8\\
\midrule
meta:gcca \hfill (hyperparams on dev set) & $d=1024$&{\bftab 72.8}/{\bftab 71.6}&{\bftab 69.6}/{\bftab 69.4}&{\bftab 81.7}/{\bftab 79.5}&{\bftab 84.2}/{\bftab 85.5}&81.3/83.3&{\bftab 83.9}/{\bftab 84.4}\\
meta:ae \hfill (hyperparams on dev set) & $d=1024$ & 71.5/70.6&68.5/68.4&80.1/78.5&82.5/83.1&80.4/81.9&82.1/83.3\\
\midrule
\midrule
\multicolumn{2}{l|}{\citet{ethayarajh2018unsupervised} \hfill (unsupervised)} & 68.3/-\blk & 66.1/-\blk & 78.4/-\blk & 79.0/-\blk & \blk-/-\blk & 79.5/-\blk \\ 
\multicolumn{2}{l|}{\citet{wieting2018paranmt} \hfill (unsupervised)} & 68.0/-\blk & 62.8/-\blk & 77.5/-\blk & 80.3/-\blk & 78.3/-\blk & 79.9/-\blk \\
\multicolumn{2}{l|}{\citet{tang2019improving} \hfill (unsupervised meta)} & 64.0/-\blk & 61.7/-\blk & 73.7/-\blk & 77.2/-\blk & 76.7/-\blk & -\blk \\
\multicolumn{2}{l|}{\citet{hassan2019uests}$^\dagger$ \hfill (unsupervised meta)} & 67.7/-\blk & 64.6/-\blk & 75.6/-\blk & 80.3/-\blk & 79.3/-\blk & 77.7/-\blk \\
\multicolumn{2}{l|}{\citet{poerner2019multiview} \hfill (unsupervised meta)} &\blk-/-\blk&\blk-/-\blk&\blk-/-\blk&\blk-/-\blk&\blk-/-\blk& 80.4/-\blk \\
\midrule
\multicolumn{2}{l|}{\citet{reimers2019sentence} \hfill (sup. siamese SoTA)} &\blk-/-\blk&\blk-/-\blk&\blk-/-\blk&\blk-/-\blk&\blk-/-\blk& \blk-/86.2 \\
\multicolumn{2}{l|}{\citet{raffel2019exploring} \hfill (supervised SoTA)} &\blk-/-\blk&\blk-/-\blk&\blk-/-\blk&\blk-/-\blk&\blk-/-\blk& 93.1/92.8 \\
\bottomrule
\end{tabularx}
\caption{Results on STS12--16 and STS Benchmark test set. STS12--16: mean Pearson's $r \times100$ / Spearman's $\rho \times 100$. STS Benchmark: overall Pearson's $r \times100$ / Spearman's $\rho \times100$. Evaluated by SentEval \cite{conneau2018senteval}. \textbf{Boldface:} best in column (except supervised). \underline{Underlined:} best single-source method. ${}^\ast$Results for up-projections are averaged over 10 random seeds. ${}^\dagger$Unweighted average computed from \citet[Table 8]{hassan2019uests}. There is no supervised SoTA on STS12--16, as they are unsupervised benchmarks.}
\label{tab:mainresults}
\end{table*}

\section{Experiments}
\subsection{Data}
We train on all sentences of length $<60$ from the first file (\textit{news.en-00001-of-00100}) of the tokenized, lowercased Billion Word Corpus (BWC) \cite{chelba2014one} ($\sim$302K sentences).
We evaluate on STS12 -- STS16 \cite{agirre2016semeval} and the unsupervised STS Benchmark test set \cite{cer2017semeval}.\footnote{We use SentEval for evaluation \cite{conneau2018senteval}. Since original SentEval does not support the unsupervised STS Benchmark, we use an implementation by \citet{singh2019gloss} (\url{https://github.com/sidak/SentEval}). We manually add the missing STS13-SMT subtask.}
These datasets consist of triples $(s_1, s_2, y)$, where $s_1, s_2$ are sentences and $y$ is their ground truth semantic similarity.
The task is to predict similarity scores $\hat{y}$ that correlate well with $y$.
We predict $\hat{y} = \mathrm{cos}(\mathcal{F}(s_1), \mathcal{F}(s_2))$.

\subsection{Metrics}
Previous work on STS differs with respect to (a) the correlation metric and (b) how to aggregate the sub-testsets of STS12--16.
To maximize comparability, we report both Pearson's $r$ and Spearman's $\rho$.
On STS12--16, we aggregate by a non-weighted average, which diverges from the original shared tasks \cite{agirre2016semeval} but ensures comparability with more recent baselines \cite{wieting2018paranmt, ethayarajh2018unsupervised}.
Results for individual STS12--16 sub-testsets can be found in the Appendix.

\subsection{Ensemble}
We select our ensemble according to the following criteria:
Every encoder should have near-SoTA performance on the unsupervised STS benchmark, and the encoders should not be too similar with regards to their training regime.
For instance, we do not use \citet{ethayarajh2018unsupervised}, which is a near-SoTA unsupervised method that uses the same word vectors as ParaNMT (see below).

We choose the Universal Sentence Encoder (USE)\footnote{\url{https://tfhub.dev/google/universal-sentence-encoder/2}} \cite{cer2018universal}, which is a Transformer trained on skip-thought, conversation response prediction and Natural Language Inference (NLI), Sentence-BERT (SBERT)\footnote{\url{https://github.com/UKPLab/sentence-transformers}. We use the \textit{large-nli-mean-tokens} model, which was \textbf{not} finetuned on STS.} \cite{reimers2019sentence}, which is a pre-trained BERT transformer finetuned on NLI, and ParaNMT\footnote{\url{https://github.com/jwieting/para-nmt-50m}} \cite{wieting2018paranmt}, which averages word and 3-gram vectors trained on backtranslated similar sentence pairs.
To our knowledge, ParaNMT is the current single-source SoTA on the unsupervised STS Benchmark.

\subsection{Hyperparameters}
We set $d=1024$ in all experiments, which corresponds to the size of the biggest single-source embedding (SBERT).
The value of $\tau$ (GCCA), as well as the autoencoder depth and loss function are tuned on the STS Benchmark development set (see Table \ref{tab:hyper}).
We train the autoencoder for a fixed number of 500 epochs with a batch size of 10,000.
We use the Adam optimizer \cite{kingma2014adam} with $\beta_1=0.9$, $\beta_2=0.999$ and learning rate $0.001$.

\subsection{Baselines}
Our main baselines are our single-source embeddings.
\citet{wieting2019no} warn that high-dimensional sentence representations can have an advantage over low-dimensional ones, i.e., our meta-embeddings might be better than lower-dimensional single-source embeddings due to size alone.
To exclude this possibility, we also up-project smaller embeddings by a random $d \times d_j$ matrix sampled from:
$$
\mathcal{U}(-\frac{1}{\sqrt{d_j}}, \frac{1}{\sqrt{d_j}})
$$
Since the up-projected sentence embeddings perform slightly worse than their originals (see Table \ref{tab:mainresults}, rows 4--5), we are confident that performance gains by our meta-embeddings are due to content rather than size.

\begin{table}
\footnotesize
\narrow
\centering
\begin{tabularx}{.98\columnwidth}{X|c|ccc}
\toprule
& full & without & without & without \\
& ensemble & ParaNMT & USE & SBERT \\
\midrule
meta:svd & {\bftab 85.0}/{\bftab 85.4} & 79.6/81.3 & 79.7/81.4 & 83.7/83.5 \\
meta:gcca & {\bftab 85.5}/{\bftab 86.1} & 84.9/84.8 & 83.8/83.8 & 85.4/85.4 \\
meta:ae & {\bftab 85.1}/{\bftab 85.5} & 76.5/80.3 & 82.5/83.5 & 28.7/41.0  \\
\bottomrule
\end{tabularx}
\caption{Ablation study: Pearson's $r \times 100$ / Spearman's $\rho \times 100$ on STS Benchmark development set when one encoder is left out.}

\label{tab:ablation}
\end{table} 

\subsection{Results}
Table \ref{tab:mainresults} shows that even the worst of our meta-embeddings consistently outperform their single-source components.
This underlines the overall usefulness of ensembling sentence encoders, irrespective of the method used.

GCCA outperforms the other meta-embeddings on five out of six datasets.
We set a new unsupervised SoTA on the unsupervised STS Benchmark test set, reducing the gap with the supervised siamese SoTA of \citet{reimers2019sentence} from 7\% to 2\% Spearman's $\rho$.

Interestingly, the naive meta-embedding methods
(concatenation and averaging)
are competitive with SVD and the autoencoder, despite not needing any unsupervised training.
In the case of concatenation, this comes at the cost of increased dimensionality, which may be problematic for downstream applications.
The naive averaging method by \citet{coates2018frustratingly} however does not have this problem, while performing only marginally worse than concatenation.

\subsection{Ablation}
Table \ref{tab:ablation} shows that all single-source embeddings contribute positively to the meta-embeddings, which supports their hypothesized complementarity.
This result also suggests that further improvements may be possible by extending the ensemble.

\subsection{Computational cost}
\subsubsection{Training}
All of our meta-embeddings are fast to train, either because they have closed-form solutions (GCCA and SVD) or because they are lightweight feed-forward nets (autoencoder).
The underlying sentence encoders are more complex and slow, but since we do not update them, we can apply them to the unlabeled training data once and then reuse the results as needed.

\subsubsection{Inference}
As noted in Section \ref{sec:relatedsts}, cross-sentence attention systems do not scale well to many real-world STS-type tasks, as they do not allow individual sentence embeddings to be cached.
Like \citet{reimers2019sentence}, our meta-embeddings do not have this problem.
This should make them more suitable for tasks like sentence clustering or real-time semantic search.

\section{Conclusion}
Inspired by the success of word meta-embeddings, we have shown how to apply different meta-embedding techniques to ensembles of sentence encoders.
All sentence meta-embeddings consistently outperform their individual single-source components on the STS Benchmark and the STS12--16 datasets, with a new unsupervised SoTA set by our GCCA meta-embeddings.
Because sentence meta-embeddings are agnostic to the size and specifics of their ensemble, it should be possible to add new encoders to the ensemble, potentially improving performance further.

\paragraph{Acknowledgments.}
This work was supported
by Siemens AG and by the European Research Council (\# 740516).

\bibliography{main}
\bibliographystyle{acl_natbib}


\appendix
\begin{table*}
\footnotesize
\verynarrow
\centering
\begin{tabularx}{.99\textwidth}{X|ccc|ccccc}
\multicolumn{1}{c}{} & \multicolumn{3}{c}{single-source embeddings} & \multicolumn{5}{c}{meta-embeddings} \\
\toprule
method: & ParaNMT & SBERT &   USE &    conc &     avg &     svd &    gcca &      ae \\
dimensionality: & $d=600$ & $d=1024$ & $d=512$ & $d=2136$ & $d=1024$ & $d=1024$ & $d=1024$ & $d=1024$ \\
\midrule
STS12 \\
\midrule
MSRpar            &    55.25/55.15 &  58.11/60.42 &  34.05/39.24 &  60.13/60.53 &  58.90/59.71 &  59.56/60.24 &  {\bftab 62.79}/{\bftab 63.90} &  61.64/63.57 \\
MSRvid            &    88.53/88.48 &  87.93/89.73 &  89.46/90.75 &  {\bftab 91.51}/92.16 &  91.29/91.92 &  91.28/91.98 &  91.20/{\bftab 92.29} &  90.69/91.69 \\
SMTeuroparl       &    53.15/59.31 &  59.63/62.40 &  49.00/62.08 &  58.99/64.02 &  {\bftab 60.16}/{\bftab 64.73} &  57.03/62.17 &  56.40/61.23 &  55.13/60.14 \\
OnWN     &    73.42/69.82 &  68.08/68.51 &  71.66/65.81 &  77.89/73.05 &  77.53/73.00 &  77.80/73.12 &  {\bftab 77.90}/{\bftab 73.50} &  75.35/73.03 \\
SMTnews  &    67.03/58.53 &  60.75/53.11 &  68.66/61.29 &  74.85/66.53 &  74.54/66.88 &  73.73/66.48 &  {\bftab 75.75}/{\bftab 67.31} &  74.91/64.76 \\
\midrule
STS13 \\
\midrule
FNWN              &    53.01/54.44 &  57.06/57.22 &  48.07/49.34 &  {\bftab 64.11}/{\bftab 64.91} &  63.46/64.26 &  63.28/63.49 &  62.74/63.54 &  63.99/64.61 \\
OnWN              &    75.62/75.80 &  77.54/80.00 &  66.64/68.10 &  80.84/81.13 &  80.46/80.81 &  79.89/80.53 &  {\bftab 84.04}/{\bftab 83.65} &  80.17/81.50 \\
SMT               &    42.54/41.13 &  44.54/44.80 &  43.85/41.80 &  47.46/44.89 &  47.87/45.04 &  48.59/45.58 &  {\bftab 49.20}/{\bftab 46.01} &  48.92/45.40 \\
headlines         &    79.52/79.83 &  73.67/77.17 &  70.70/71.82 &  81.13/83.48 &  80.64/82.96 &  81.49/83.54 &  {\bftab 82.58}/{\bftab 84.37} &  80.78/82.13 \\
\midrule
STS14 \\
\midrule
OnWN              &    82.22/83.20 &  81.51/82.99 &  74.61/76.01 &  85.08/ 85.83 &  85.12/85.84 &  84.23/85.17 &  {\bftab 87.34}/{\bftab 87.27} &  84.24/85.09 \\
deft-forum        &    60.01/59.49 &  57.66/60.45 &  50.12/49.43 &  67.57/66.84 &  67.09/66.19 &  66.84/66.20 &  {\bftab 68.40}/{\bftab 67.26} &  67.22/66.82 \\
deft-news         &    77.46/72.75 &  72.62/76.80 &  68.35/63.35 &  {\bftab 81.72}/79.04 &  81.60/78.98 &  80.36/78.31 &  81.09/{\bftab 79.20} &  79.59/78.83 \\
headlines         &    78.85/76.98 &  73.72/75.41 &  65.88/62.34 &  79.64/79.93 &  79.39/79.86 &  79.85/79.59 &  {\bftab 81.68}/{\bftab 81.50} &  80.13/79.77 \\
images            &    86.14/83.36 &  84.57/79.42 &  85.54/80.55 &  {\bftab 89.52}/{\bftab 85.68} &  89.35/85.51 &  89.29/85.37 &  88.83/84.83 &  87.64/83.42 \\
tweet-news        &    79.39/73.43 &  75.12/70.80 &  72.48/64.24 &  82.50/76.50 &  82.12/76.13 &  {\bftab 83.14}/77.17 &  83.09/77.04 &  81.61/{\bftab 77.23} \\
\midrule
STS15 \\
\midrule
answers-forums    &    73.54/74.50 &  64.04/62.78 &  72.70/75.02 &  {\bftab 79.33}/{\bftab 79.91} &  78.47/79.12 &  79.15/79.69 &  78.39/78.59 &  72.65/72.21 \\
answers-stud.  &    77.06/77.87 &  79.12/80.14 &  60.99/63.32 &  81.01/82.10 &  80.15/81.45 &  81.02/82.14 &  80.86/82.18 & {\bftab 83.03}/{\bftab 83.56} \\
belief            &    80.28/80.25 &  77.46/77.46 &  78.68/82.14 &  86.14/{\bftab 87.58} &  85.55/87.01 &  85.05/86.02 & {\bftab 86.38}/{\bftab 87.58} &  82.49/83.07 \\
headlines         &    81.92/82.28 &  78.91/81.88 &  73.26/74.77 &  83.20/86.03 &  83.33/86.25 &  83.48/86.02 &  {\bftab 84.87}/{\bftab 86.72} &  84.16/85.53 \\
images            &    88.60/88.87 &  86.76/89.02 &  88.39/90.34 &  {\bftab 90.92}/{\bftab 91.95} &  90.86/91.92 &  90.46/91.59 &  90.34/91.85 &  90.26/91.35 \\
\midrule
STS16 \\
\midrule
answer-answer     &    69.71/68.96 &  63.41/66.63 &  72.52/72.72 &  {\bftab 79.65}/78.89 &  78.93/77.82 &  79.37/{\bftab 79.21} &  78.70/78.50 &  76.83/77.17 \\
headlines         &    80.47/81.90 &  75.23/79.33 &  69.70/75.11 &  80.97/{\bftab 84.95} &  80.60/84.53 &  81.36/85.14 &  {\bftab 81.41}/84.85 &  80.40/83.17 \\
plagiarism        &    84.49/85.62 &  80.78/82.04 &  74.93/77.42 &  85.86/87.17 &  85.88/87.25 &  85.54/87.36 &  {\bftab 85.92}/{\bftab 87.76} &  85.01/86.14 \\
postediting       &    84.53/86.34 &  81.32/85.87 &  82.81/86.49 &  {\bftab 88.18}/{\bftab 90.76} &  87.98/90.51 &  87.55/90.21 &  87.01/90.24 &  86.71/89.28 \\
question-quest. &    72.37/72.73 &  63.38/64.72 &  68.54/70.25 &  75.49/77.42 &  {\bftab 76.05}/{\bftab 77.76} &  74.08/75.93 &  73.44/74.98 &  73.25/73.60 \\
\bottomrule
\end{tabularx}
\caption{Pearson's $r$ / Spearman's $\rho$ $\times 100$ on individual sub-testsets of STS12--STS16. \textbf{Boldface}: best method in row.}
\end{table*}

\end{document}